\documentclass{article}

\usepackage{arxiv}

\usepackage[utf8]{inputenc}
\usepackage[T1]{fontenc}

\usepackage{amsmath,amssymb,amsfonts}

\usepackage{booktabs}
\usepackage{array}
\usepackage{multirow}
\usepackage{makecell}
\usepackage{tabularx}
\usepackage[table]{xcolor}

\usepackage{graphicx}
\graphicspath{{./images/}}

\usepackage[hidelinks]{hyperref}


\usepackage[final]{microtype}

\definecolor{gold}{RGB}{255,215,0}
\definecolor{silver}{RGB}{192,192,192}
\definecolor{bestoverall}{RGB}{144,238,144}
\definecolor{secondbest}{RGB}{255,248,220}

\title{Less is More: Strategic Expert Selection Outperforms Ensemble Complexity in Traffic Forecasting}

\author{
  Walid Guettala \\
  Department of Artificial Intelligence \\
  ELTE Eötvös Loránd University \\
  Budapest, Hungary \\
  \texttt{guettalawalid@inf.elte.hu}
  \And
  Yufan Zhao \\
  Department of Artificial Intelligence \\
  ELTE Eötvös Loránd University \\
  Budapest, Hungary \\
  \texttt{a93zfp@inf.elte.hu}
  \And
  László Gulyás \\
  Department of Artificial Intelligence \\
  ELTE Eötvös Loránd University \\
  Budapest, Hungary \\
  \texttt{lgulyas@inf.elte.hu}
}

\date{\vspace{-0.5em}\small Preprint. Accepted to IEEE ICTAI 2025. This manuscript differs from the published version in formatting and minor wording.}

\begin{document}
\maketitle

\begin{abstract}
Traffic forecasting is fundamental to intelligent transportation systems, enabling congestion mitigation and emission reduction in increasingly complex urban environments. While recent graph neural network approaches have advanced spatial-temporal modeling, existing mixture-of-experts frameworks like Time-Enhanced Spatio-Temporal Attention Model (TESTAM) lack explicit incorporation of physical road network topology, limiting their spatial capabilities. We present \textbf{TESTAM+}, an enhanced spatio-temporal forecasting framework that introduces a novel \textit{SpatioSemantic Expert} integrating physical road topology with data-driven feature similarity through hybrid graph construction. TESTAM+ achieves significant improvements over TESTAM: 1.3\% MAE reduction on METR-LA (3.10 vs.\ 3.14) and 4.1\% improvement on PEMS-BAY (1.65 vs.\ 1.72). Through comprehensive ablation studies, we discover that strategic expert selection fundamentally outperforms naive ensemble aggregation. Individual experts demonstrate remarkable effectiveness: the \textit{Adaptive Expert} achieves 1.63 MAE on PEMS-BAY, outperforming the original three-expert TESTAM (1.72 MAE), while the \textit{SpatioSemantic Expert} matches this performance with identical 1.63 MAE. The optimal Identity + Adaptive configuration achieves an 11.5\% MAE reduction compared to state-of-the-art MegaCRN on METR-LA (2.99 vs. 3.38), while reducing inference latency by 53.1\% compared to the full four-expert TESTAM+. Our findings reveal that fewer, strategically designed experts outperform complex multi-expert ensembles, establishing new state-of-the-art performance with superior computational efficiency for real-time deployment.
\end{abstract}


\section{Introduction}

Traffic forecasting is pivotal for intelligent transportation systems, supporting congestion mitigation, route optimization and emission reduction. Reliable predictions improve route planning, shorten travel times and raise transportation efficiency in complex urban settings.

Existing solutions struggle with spatial adaptability. The fundamental challenge in traffic forecasting lies in capturing the intricate spatio-temporal dependencies within road networks. Traffic patterns exhibit both recurring behaviors (e.g., daily rush hours) and non-recurring events (e.g., accidents, weather disruptions), requiring models that can adapt to diverse traffic scenarios. Traditional approaches such as auto-regressive integrated moving average (ARIMA) and support vector regression (SVR) struggled with nonlinear traffic dynamics, while early deep learning methods like recurrent neural networks (RNNs) and convolution neural networks (CNNs) inadequately captured spatial correlations in irregular road networks.

Recent advances in graph neural networks (GNNs) have significantly improved traffic forecasting by modeling road networks as graphs, where sensors represent nodes and connectivity represents edges. While graph-based models like Diffusion Convolution Recurrent Neural Network (DCRNN)~\cite{li2017diffusion} and Graph WaveNet~\cite{wu2019graph} leverage graph convolution to capture spatial dependencies and introduced learnable adjacency matrices for dynamic spatial modeling, they rely on fixed aggregation schemes that prove suboptimal across diverse traffic conditions. Recent mixture-of-experts (MoE) approaches like Time‑Enhanced Spatio‑Temporal Attention Model (TESTAM)~\cite{lee2024testam} addressed this limitation through a framework with three specialized experts, but they neglect explicit road network semantics, limiting their ability to leverage structural priors during spatial modeling.

This work presents \textbf{TESTAM+}, an enhanced spatio-temporal forecasting framework that significantly improves enhances spatial representation capacity through:
\begin{enumerate}
    \item A novel \textit{SpatioSemantic Expert} integrating physical road topology with data-driven feature similarity via hybrid graph construction, enabling the model to exploit both structural priors and adaptive traffic patterns;
    \item Dynamic expert routing that selects optimal spatial modeling strategies per context, using sparsity-controlled graph construction to reduce computational overhead;
    \item State-of-the-art performance with strategic expert selection: Individual experts demonstrate remarkable effectiveness with the \textit{Adaptive Expert} achieving 1.63 MAE on PEMS-BAY and \textit{SpatioSemantic Expert} matching this performance, both outperforming the original three-expert TESTAM (1.72 MAE). The optimal Identity + Adaptive configuration achieves 11.5\% MAE reduction on METR-LA (2.99 vs. 3.38 MegaCRN), while TESTAM+ delivers consistent improvements over TESTAM with 1.3\% and 4.1\% MAE reductions on METR-LA and PEMS-BAY respectively; 
    \item Superior computational efficiency: Strategic expert selection reduces inference latency by 53.1\% and 61.7\% compared to full four-expert configurations, demonstrating that fewer, well-designed experts outperform complex multi-expert ensembles for real-time deployment.

\end{enumerate}

This work is organized as follows. Section~\ref{sec:related} reviews related work in traffic forecasting and mixture-of-experts frameworks. Section~\ref{sec:methodology} presents our methodology, detailing the \textit{SpatioSemantic Expert} design and the overall TESTAM+ architecture. Section~\ref{sec:experiments} provides experimental results and analysis on real-world datasets. Section~\ref{sec:conclusion} concludes and outlines future research directions.

\section{Related Work \label{sec:related}}

\subsection*{Traffic Forecasting}

Early traffic prediction relied on ARIMA~\cite{vlahogianni2014short} and support vector regression~\cite{li2018brief}, which assumed linear patterns but struggled with nonlinear traffic dynamics. RNNs, particularly Long Short-Term Memory (LSTM)~\cite{hochreiter1997long} variants, improved temporal modeling but inadequately handled spatial correlations. Hybrid CNN-RNN architectures~\cite{yu2017spatio} addressed this by using convolution layers for spatial structure and recurrent components for temporal dynamics.

Graph-based methods emerged for irregular road networks. DCRNN~\cite{li2017diffusion} and Graph WaveNet~\cite{wu2019graph} leverage graph convolution networks to capture spatial dependencies by modeling sensors as nodes and connectivity as edges. Recent advances include dynamic graph construction through attention mechanisms, such as the Attention-based Spatial-Temporal Graph Convolution Network (ASTGCN)~\cite{guo2019attention}, and adaptive adjacency learning in the Dynamic Graph Convolution Recurrent Network (DGCRN)~\cite{li2023dynamic}. 

Two-stream approaches like the Spatio-Temporal Multi-Graph Convolution Network (ST-MGCN)~\cite{9288187} decompose traffic into stable and dynamic components, while attention-based methods such as the Spatio-Temporal Multi-Graph Attention Network (ST-MGAT)~\cite{9288309} leverage multi-head attention mechanisms to capture dynamic spatial relationships, though these approaches still rely on fixed aggregation strategies across diverse traffic conditions.

Existing approaches rely on fixed spatial aggregation schemes, proving suboptimal across diverse traffic scenarios. Our work addresses this through context-aware expert selection that dynamically adapts spatial modeling strategies.

\subsection*{Spatio-Temporal Modeling}

Early approaches adopted decoupled designs with separate spatial and temporal modules (Spatio-Temporal Graph Convolution Network (ST-GCN))~\cite{yu2017spatio}, Graph WaveNet~\cite{wu2019graph}). Recent unified architectures attempt joint modeling through spectral fusion (Spectral Temporal Graph Neural Network (StemGNN))~\cite{cao2020spectral} and attention mechanisms (Graph Multi-Attention Network (GMAN))~\cite{zheng2020gman}. However, these maintain static designs that assume consistent strategies across inputs, limiting performance under variable traffic conditions.

Our framework enables dynamic routing among multiple spatial modeling strategies, enhancing generalization and responsiveness.

\subsection*{Mixture of Experts}

MoE frameworks~\cite{shazeer2017outrageously} scale model capacity through selective expert activation, enabling adaptive computation allocation. Multiple specialized sub-networks process inputs in parallel, allowing experts to specialize in distinct patterns. Applications span natural language processing~\cite{lepikhin2020gshard}, computer vision~\cite{riquelme2021scaling}, and time-series forecasting~\cite{sen2019think}.

TESTAM~\cite{lee2024testam} introduced mixture of experts for traffic prediction with three temporal experts: Identity (pure temporal modeling), Adaptive (learned static adjacency), and Attention (dynamic attention-based graphs). While TESTAM captures diverse temporal behaviors, it lacks explicit incorporation of physical road network topology, limiting its ability to leverage structured spatial relationships inherent in transportation networks.

Our TESTAM+ addresses this limitation by introducing the \textit{SpatioSemantic Expert}, which integrates both physical topology and data-driven similarity through hybrid graph construction. This expert combines spatial distance encoding with feature similarity computation, enabling the model to exploit both structural road network priors and adaptive traffic patterns. Unlike TESTAM's purely data-driven spatial modeling, our approach explicitly incorporates domain knowledge while maintaining adaptability, resulting in superior performance with reduced computational overhead compared to the full four-expert configuration.

\section{Methods \label{sec:methodology}}
We propose \textbf{TESTAM+}, an enhanced spatio-temporal forecasting framework that extends the original TESTAM  \cite{lee2024testam} architecture by introducing the \textit{SpatioSemantic Expert}, a novel module designed to explicitly incorporate spatial structural priors while retaining adaptability to dynamic traffic patterns. 

\subsection{Preliminaries}
\subsubsection{Problem Definition}

Let \( G = (V, A) \) be a road network with \(|V| = N\) nodes and adjacency matrix \( A \in \mathbf{R}^{N \times N} \). Stack node features at time \( t \) as \( \mathbf{X}^{(t)} \in \mathbf{R}^{N \times C} \), where \( C \) is the number of input features per node (e.g., speed, flow, occupancy). Given a sequence of past observations
\[
\bigl[\mathbf{X}^{(t-T'+1)}, \dots, \mathbf{X}^{(t)}\bigr] \in \mathbf{R}^{T' \times N \times C},
\]
the goal is to learn a mapping
\[
f: \mathbf{R}^{T' \times N \times C} \to \mathbf{R}^{T \times N \times C}
\]
that predicts the future node features over the next \( T \) steps:
\[
\bigl[\mathbf{X}^{(t-T'+1)}, \dots, \mathbf{X}^{(t)}\bigr] 
\xrightarrow{f} 
\bigl[\mathbf{X}^{(t+1)}, \dots, \mathbf{X}^{(t+T)}\bigr].
\]
Where the input consists of \( T' \) past frames and the output consists of \( T \) future frames.

\subsubsection{Time-Enhanced Spatio-Temporal Attention Model (TESTAM)} 
TESTAM is a mixture-of-experts model for traffic forecasting that dynamically combines three specialized spatio-temporal learners to capture both recurring and non-recurring patterns \cite{lee2024testam}. Input features are first augmented with Time2Vec embeddings to encode periodic and linear temporal information. These enriched features are then processed in parallel by the \textit{Identity Expert} (which ignores spatial structure and models purely temporal dependencies), the \textit{Adaptive Expert} (which learns a static adjacency matrix to exploit recurring spatial correlations), and the \textit{Attention Expert} (which constructs a fully dynamic, attention-based graph to capture sudden, non-recurring events).

Each expert follows a shared computational backbone—its own spatial module, temporal self-attention, a time-enhanced attention mechanism that maps historical observations directly to future predictions, and a point-wise feedforward layer—but differs only in the form of spatial modeling it employs. A memory-based gating network computes similarity scores between a pooled summary of the input and trainable memory keys, thereby routing each instance to the most suitable expert or combination of experts. This design prevents error accumulation typical of auto-regressive decoders and enables full parallelism in forecasting.

\subsubsection{Overall Framework}
Fig.\ref{fig:testam_plus_pipeline} shows that TESTAM+ follows a three-stage pipeline for spatio-temporal traffic forecasting:

\begin{figure}[htbp]
  \centering
  \includegraphics[width=0.7\textwidth]{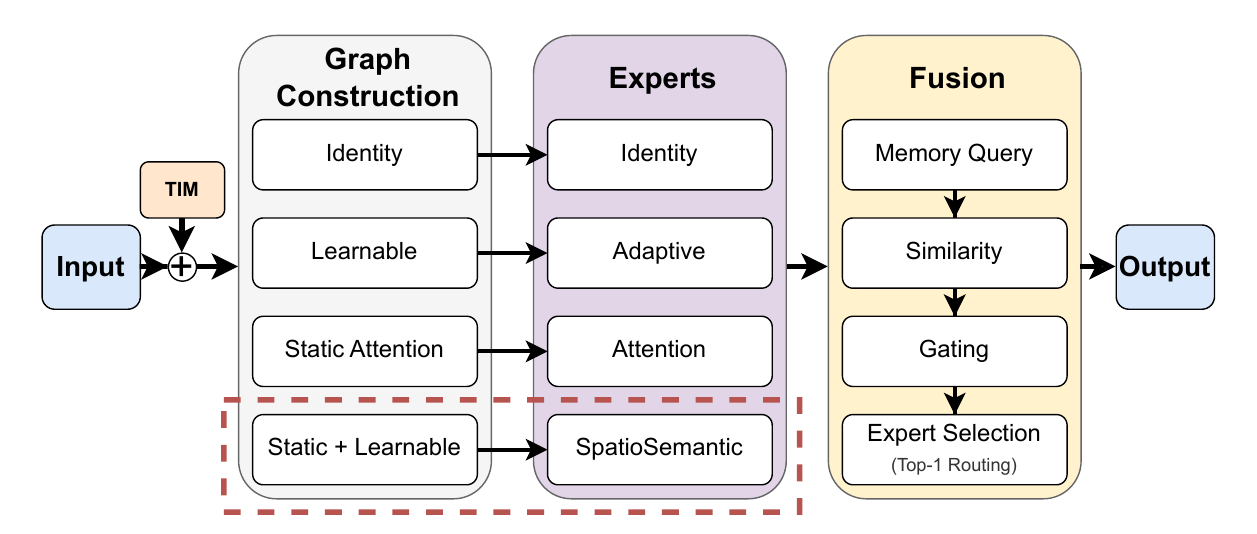}
  \caption{Overview of the TESTAM+ pipeline. Input features are first augmented with TIM and then fed in parallel to four experts: \textbf{Identity} (pure temporal modeling, no spatial structure), \textbf{Adaptive} (learned graph), \textbf{Attention} (attention-based graph), and \textbf{SpatioSemantic} (hybrid static and learnable spatial priors). Each expert constructs its own spatial graph and extracts spatio-temporal features. A memory-query similarity gating module then computes dot-product scores between a pooled input summary and trainable expert keys to perform top-1 routing, and the selected expert’s output is fused to produce the traffic forecast. The red dashed box marks the \textit{SpatioSemantic Expert} added in TESTAM+, while without it, the figure represents the original TESTAM.}
  \label{fig:testam_plus_pipeline}

\end{figure}

\begin{enumerate}
    \item[\textbf{a)}] \textbf{Graph Construction:} Each expert constructs or utilizes a different form of spatial graph, ranging from static topological graphs to learnable dynamic graphs that encode spatial similarity and connectivity.
    \item[\textbf{b)}] \textbf{Experts:} Then each expert independently performs spatio-temporal feature extraction using a shared structure that includes temporal attention, graph convolutions, and feedforward transformations.
    \item[\textbf{c)}] \textbf{Fusion:} Expert outputs are combined through a gating mechanism that performs context-aware routing based on input characteristics.
\end{enumerate}

This modular design allows TESTAM+ to incorporate both prior knowledge and data-driven patterns, enhancing generalization and adaptability.

\subsubsection{Temporal Information Embedding (TIM)}

TESTAM+ replaces the Transformer’s positional encoding with a Time2Vec embedding~\cite{time2vec} to capture both linear and periodic components of each scalar timestamp \( v(\tau) \), where \( \tau \) denotes the time index. The embedding is defined as
\begin{align}
\mathrm{TIM}_i(\tau) &= G_i(w_i\,v_i(\tau) + \phi_i) \\
G_i &= 
\begin{cases}
\mathrm{Identity}, & \text{if } i = 0 \\
F, & \text{otherwise}
\end{cases}
\end{align}
where each \( w_i \) and \( \phi_i \) is learnable, and \( F \) is a periodic activation function (e.g., \( \sin \)). The resulting \( \mathrm{TIM}(\tau) \) is concatenated with the original feature vector \( \mathbf{x}(\tau) \), and a linear projection is applied:
\begin{equation}
\mathbf{x}'(\tau) = W_{\mathrm{proj}}\bigl[\mathbf{x}(\tau) \,\|\, \mathrm{TIM}(\tau)\bigr]
\end{equation}
to align the combined representation with the model’s hidden dimension. This allows the model to encode both linear and periodic temporal patterns directly. The resulting vector \( \mathbf{x}'(\tau) \) is then fed to the expert models.

\subsection{Expert Models}
\subsubsection{Shared Model Structure}
All experts in TESTAM+ share a modified Transformer backbone for feature learning, consisting of four blocks: Temporal Attention, Spatial Model, Time-Enhanced Attention, and Feedforward. Each block applies a skip connection using residual summation followed by layer normalization. Given input features $\mathbf{X}' \in \mathbf{R}^{N \times T \times d}$, each expert performs the following sequence:

\begin{enumerate}
    \item[\textbf{a)}] \textbf{Temporal Attention:} For each node, a self-attention mechanism captures dependencies across the time axis:
    \begin{equation}
    \mathbf{H}_t^{(i)} = \text{Attention}(Q^{(i)}, K^{(i)}, V^{(i)}), \quad \forall i \in \{1, \ldots, N\}
    \end{equation}
    where $Q$, $K$, and $V$ are the query, key, and value projections of the input.
    
    \item[\textbf{b)}] \textbf{Spatial Model:} Each node aggregates spatially neighboring features using a predefined or learned adjacency matrix $\hat{A}$, depending on the expert type:
    \begin{equation}
    \mathbf{H}_s = \sigma\left( \hat{A} \mathbf{X} W_s \right)
    \end{equation}
    where $W_s$ is a learnable weight and $\sigma(\cdot)$ is a non-linear activation.

    \item[\textbf{c)}] \textbf{Time-Enhanced Attention:} This block models direct dependencies between historical and future time steps using cross-attention. For each target time step $j$, attention is computed between input features $H^{(i)}$ and the Time2Vec embedding $\mathrm{TIM}(\tau^{(j)})$ as:
    \begin{equation}
    e_{i,j} = \frac{(H^{(i)} W_q)(\mathrm{TIM}(\tau^{(j)}) W_k)^\top}{\sqrt{d_k}}
    \label{eq:teattn_score}
    \end{equation}
    \begin{equation}
    \alpha_{i,j} = \frac{\exp(e_{i,j})}{\sum_{k} \exp(e_{i,k})}
    \label{eq:teattn_weight}
    \end{equation}
    This enables the model to generate future representations in parallel, eliminating auto-regressive dependencies.

    \item[\textbf{d)}] \textbf{Feedforward Layer:} The concatenated or residual output is passed through a position-wise feedforward network to enhance non-linearity and representation capacity:
    \begin{equation}
    \mathbf{H}_f = \text{FFN}\left(\mathbf{H}_t + \mathbf{H}_s\right)
    \end{equation}
\end{enumerate}

This structure ensures consistent computation across all experts while allowing their internal graphs to capture diverse spatial semantics.

\subsubsection{Per-experts}

The experts in TESTAM+ each offer unique spatio-temporal methods. While they follow the shared structure outlined previously, they differ in the type of graph they utilize and the specific focus of their feature extraction in the spatial model block (Fig.\ref{fig:experts_description}). The experts include:

\begin{figure}[htbp]
  \centering
  \includegraphics[width=0.34\textwidth]{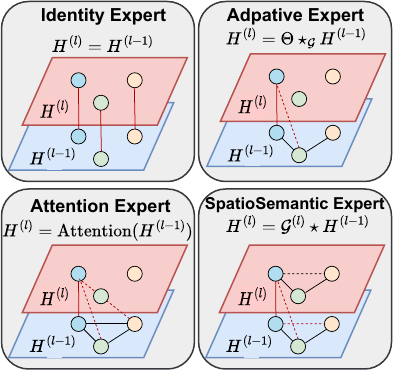}
  \caption{The four expert spatial model block in TESTAM+. Black lines denote spatial connectivity and red lines indicate information flow along those connections. The \textit{Identity Expert} focuses solely on temporal dependencies without spatial edges. The \textit{Adaptive Expert} learns a static graph to capture recurring spatial relations. The \textit{Attention Expert} dynamically infers spatial connectivity via attention to model non-recurring interactions. The \textit{SpatioSemantic Expert} integrates physical road topology and data-driven similarity to construct a hybrid spatial graph.}
  \label{fig:experts_description}
\end{figure}

\begin{itemize}
    \item \textbf{\textit{Identity Expert}:} Models temporal relationships based on a predefined identity adjacency matrix that reflects the fixed network topology.
    \item \textbf{\textit{Adaptive Expert}:} Uses a learned adjacency matrix to dynamically adjust the spatial and temporal graph according to the data, enabling better adaptability to traffic patterns.
    \item \textbf{\textit{Attention Expert}:} Employs an attention mechanism to focus on the most relevant spatial-temporal dependencies, leveraging both spatial and temporal attention weights for enhanced prediction accuracy.
\end{itemize}

\subsubsection{SpatioSemantic Expert}
The original TESTAM framework leverages a Mixture-of-Experts (MoE) mechanism to model heterogeneous temporal behaviors but lacks explicit semantic graph learning. This limitation restricts the model's capacity to capture structured road network interactions. To address this issue, we introduce the \textit{SpatioSemantic Expert}, which constructs a dynamic graph that integrates physical topology and data-driven similarity. The method consists of three primary stages: (1) hybrid graph construction, (2) graph normalization and sparsity control, and (3) expert-based spatio-temporal modeling.

Building on the previously defined graph \( G = (V, A) \). For each node pair $(i, j)$, we compute an adaptive edge score $e_{ij}$ via a shared multi-layer perceptron (MLP):

\begin{equation}
e_{ij} = F\left( \text{spatial\_distance}_{ij} \oplus \text{similarity}(v_i, v_j) \right)
\label{eq:spatial_expert}
\end{equation}

where $\oplus$ denotes feature concatenation, $\text{spatial\_distance}_{ij} \in \{0, 1\}$ encodes road topology (1 if there exists a direct road link), and $\text{similarity}(v_i, v_j)$ represents a data-driven feature similarity score. The function $F(\cdot)$ denotes a shared MLP applied across all node pairs.
\begin{enumerate}

\item[\textbf{a)}] \textbf{Sparsity Constraint}
To suppress noise and reduce computational complexity, we retain only the top-$k$ highest scoring edges across all node pairs. The sparsity level is determined by:

\begin{equation}
k = \left\lceil N^2 \cdot \rho \right\rceil
\label{eq:sparsity}
\end{equation}

where $N = |V|$ is the number of nodes, and $\rho \in (0, 1]$ is the sparsity ratio. In our experiments, we empirically set $\rho = 0.7$.

\item[\textbf{b)}] \textbf{Graph Normalization}
The resulting sparse adjacency matrix $A$ is row-normalized to produce a stochastic transition matrix $\hat{A}$ used in subsequent message-passing operations:

\begin{equation}
\hat{A}_{ij} = \frac{A_{ij}}{\sum_j A_{ij}}
\label{eq:normalization}
\end{equation}

This ensures that the outgoing edge weights from each node sum to one.

\item[\textbf{c)}] \textbf{Expert Module Pipeline}
Given the normalized adjacency matrix $\hat{A}$, the \textit{SpatioSemantic Expert} performs a multi-stage process:

\begin{itemize}
    \item \textbf{Temporal Attention and Time-Enhanced Attention:} Captures non-local temporal dependencies across time steps.
    \item \textbf{Graph Convolution:} Aggregates features from spatially connected neighbors using $\hat{A}$.
    \item \textbf{Feedforward Layer:} Applies a non-linear transformation to enrich representational capacity.
\end{itemize}

The final output of the \textit{SpatioSemantic Expert} is integrated into the expert mixture through the gating mechanism of TESTAM+, enabling context-aware expert selection.
\end{enumerate}

\subsection{Gating Mechanism}

\subsubsection{Memory Query}
To route each input to the most suitable expert, TESTAM+ uses a memory-based query mechanism. For an input sequence $\mathbf{X}$, we compute a summary query vector $\mathbf{q} \in \mathbf{R}^d$ by global average pooling followed by a linear projection:
\begin{equation}
\mathbf{q} = W_q \cdot \text{AvgPool}(\mathbf{X})
\end{equation}

Each expert $m$ is associated with a key vector $\mathbf{k}_m$ stored in a trainable memory bank.

\subsubsection{Routing Decision}
The routing weights are computed by a dot-product similarity followed by softmax normalization:
\begin{equation}
\alpha_m = \frac{\exp(\mathbf{q}^\top \mathbf{k}_m)}{\sum_{m'=1}^{M} \exp(\mathbf{q}^\top \mathbf{k}_{m'})}
\end{equation}
where $\alpha_m$ is the routing probability assigned to expert $m$. The final prediction is the weighted sum of expert outputs:
\[
\mathbf{y} = \sum_{m=1}^{M} \alpha_m \cdot f_m(\mathbf{X})
\]

\subsubsection{Routing Loss (Optional)}
To encourage expert specialization, an auxiliary routing classification loss can be introduced. If an oracle label $y_{\text{expert}}$ exists, the routing loss is:
\begin{equation}
\mathcal{L}_{\text{route}} = \text{CrossEntropy}(y_{\text{expert}}, \boldsymbol{\alpha})
\end{equation}
This term promotes sharper expert selection and prevents uniform averaging.

\section{Experimental Setup}

We evaluate our proposed TESTAM+ framework against the baseline TESTAM model and state-of-the-art methods to demonstrate the effectiveness of the \textit{SpatioSemantic Expert} in enhancing spatial and prediction accuracy.

\textbf{Baseline and Model Configuration:} Our baseline is the original TESTAM~\cite{lee2024testam} with three experts (Identity, Adaptive, Attention). TESTAM+ introduces the fourth \textit{SpatioSemantic Expert} while maintaining identical architectural components and training settings for fair comparison. The \textit{SpatioSemantic Expert} leverages hybrid graph construction combining physical connectivity with similarity-based dynamic relationships.

We compare against the methods listed in Table~\ref{tab:60min_results}: MTGNN \cite{wu2020connecting}, AGCRN \cite{bai2020adaptive}, CCRNN \cite{ye2021coupled}, GTS \cite{shang2021discrete}, PM-MemNet \cite{lee2110learning}, MegaCRN \cite{jiang2023spatio}, and TESTAM \cite{lee2024testam}. While LSTM baselines are commonly used, recent comprehensive evaluations \cite{jiang2023spatio} demonstrate that graph-based approaches consistently outperform sequential models by 15-20\% on traffic forecasting tasks, making GNN baselines more relevant for fair comparison.

\textbf{Datasets:} We conduct experiments on two benchmark datasets: METR-LA contains four months of traffic speed data from 207 highway sensors in Los Angeles (34,272 time steps), while PEMS-BAY includes six months from 325 sensors in the Bay Area (52,116 time steps). Both datasets provide 5-minute interval measurements with z-score normalization applied. Data is split chronologically: 70\% training, 10\% validation, 20\% testing.

\textbf{Task Formulation:} The prediction task is multi-step traffic speed forecasting using a 12-step historical window (1 hour) to predict the subsequent 12 time steps. Training employs batch size 64, dropout rate 0.2, Adam optimizer with learning rate 0.001, early stopping (patience 15, max 100 epochs).

\textbf{Evaluation Metrics:} We employ three standard metrics: Mean Absolute Error (MAE) for overall accuracy, Root Mean Squared Error (RMSE) emphasizing larger discrepancies, and Mean Absolute Percentage Error (MAPE) for scale-invariant comparison. All experiments are conducted on NVIDIA Quadro RTX 6000 GPU using PyTorch and PyTorch Geometric frameworks.

\section{Experimental Results \label{sec:experiments}}
In this section we present a thorough evaluation of our mixture-of-experts framework on two standard traffic-forecasting benchmarks. First, we report long-term (60-minute) prediction results, highlighting significant gains over state-of-the-art baselines. Next, we analyze the individual and combined expert configurations to reveal how spatial-modeling diversity and routing design contribute to both accuracy and inference efficiency. Together, these experiments demonstrate the predictive power and practical scalability of our approach.

\subsection{Long-term Traffic Prediction Performance}

We evaluate our model variants on METR-LA and PEMS-BAY datasets under a challenging 60-minute forecasting horizon, as presented in Table~\ref{tab:60min_results}. Our results demonstrate substantial improvements over existing state-of-the-art methods, with the different configurations achieving optimal performance across datasets.

\begin{table}[htbp]
\caption{Long-term (60-minute horizon) traffic forecasting performance on METR-LA and PEMS-BAY datasets. Metrics reported: Mean Absolute Error (MAE), Root Mean Squared Error (RMSE), and Mean Absolute Percentage Error (MAPE); ↓ indicates lower is better. Configurations: Id–Identity, Ad–Adaptive, SS–SpatioSemantic. Methods marked with \textbf{*} are our proposed variants. Best results are in \textbf{bold}, second-best are \underline{underlined}.}

\label{tab:60min_results}
\centering\footnotesize
\setlength{\tabcolsep}{4pt}
\begin{tabular}{@{}l ccc ccc@{}}
\toprule
\multirow{2}{*}{\textbf{Method}} & \multicolumn{3}{c}{\textbf{METR-LA}} & \multicolumn{3}{c}{\textbf{PEMS-BAY}} \\
\cmidrule(lr){2-4} \cmidrule(lr){5-7}
 & \makecell[c]{MAE} & \makecell[c]{RMSE} & \makecell[c]{MAPE} & \makecell[c]{MAE} & \makecell[c]{RMSE} & \makecell[c]{MAPE} \\
\midrule
MTGNN (\cite{wu2020connecting}, 2020) & 3.49 & 7.23 & 9.87\% & 1.94 & 4.49 & 4.53\% \\
AGCRN (\cite{wu2020connecting}, 2020) & 3.68 & 7.56 & 10.46\% & 1.98 & 4.59 & 4.63\% \\
CCRNN (\cite{ye2021coupled}, 2021) & 3.73 & 7.65 & 10.59\% & 2.07 & 4.65 & 4.87\% \\
GTS (\cite{shang2021discrete}, 2021) & 3.47 & 7.29 & 9.83\% & 1.98 & 4.56 & 4.59\% \\
PM-MemNet (\cite{lee2110learning}. 2022) & 3.46 & 7.29 & 9.97\% & 1.95 & 4.49 & 4.54\% \\
MegaCRN (\cite{jiang2023spatio}, 2023) & 3.38 & 7.23 & 9.72\% & 1.88 & 4.42 & 4.41\% \\
TESTAM (\cite{lee2024testam}, 2024) & 3.14 & 6.19 & 8.67\% & 1.72 & 3.71 & 3.91\% \\
Ad\textbf{*} & 3.11 & 6.17 & \underline{8.34\%}  & \textbf{1.63} & \textbf{3.58} & \underline{3.65}\% \\
SS\textbf{*} & 3.13 & 6.33 & 8.65\% & \textbf{1.63} & \underline{3.61} & \underline{3.65}\% \\
Id+Ad\textbf{*} & \textbf{2.99} & \textbf{6.13} & \textbf{8.11}\% & \underline{1.65} & 3.67 & \textbf{3.62}\% \\
TESTAM+\textbf{*} & \underline{3.10} & \underline{6.16} & \underline{8.34\%} & \underline{1.65} &3.65 & 3.73\% \\
\bottomrule

\end{tabular}
\end{table}

The Identity + \textit{Adaptive Expert} (Id+Ad) configuration delivers best results results on METR-LA, reducing MAE to 2.99 on METR-LA compared to the previous best baseline MegaCRN at 3.38, representing an \textbf{11.5\% improvement}. On PEMS-BAY, individual experts demonstrate remarkable effectiveness: both the Adaptive (Ad) and SpatioSemantic (SS) experts achieve identical optimal performance with 1.63 MAE,  significantly outperforming MegaCRN's 1.88 (representing a 13.3\% enhancement) and surpassing the original three-expert TESTAM's 1.72 MAE.

TESTAM+ consistently outperforms the original TESTAM across both datasets: achieving 1.3\% MAE improvement on METR-LA (3.10 vs. 3.14) and 4.1\% improvement on PEMS-BAY (1.65 vs. 1.72). This demonstrates the effectiveness of our \textit{SpatioSemantic Expert} integration and refined routing mechanisms.

A critical insight emerges: strategic expert selection proves superior to complex ensemble approaches. Single experts (Ad, SS) outperform the original three-expert TESTAM on PEMS-BAY, while simple two-expert combinations achieve state-of-the-art results on METR-LA. This challenges conventional mixture-of-experts principles and highlights the importance of expert specialization alignment with problem structure.

The superiority of our approach is particularly evident in long-term predictions where traditional methods suffer from error accumulation. The \textit{SpatioSemantic Expert}'s integration of physical topology with data-driven similarity enables effective spatial modeling, while maintaining computational efficiency compared to full multi-expert configurations.

\subsection{Expert Configuration Analysis}

Table~\ref{tab:results_combined} provides comprehensive insights into the contribution of individual experts and their combinations, revealing critical architectural design principles. The analysis spans individual experts, two-expert combinations, three-expert ensembles, and full model configurations, alongside computational efficiency metrics.

\begin{table}[htbp]
\caption{Ablation study of expert configurations on METR-LA and PEMS-BAY datasets. We report MAE, RMSE, MAPE (↓ indicates lower is better), and inference time (in seconds). Models include single experts, two- and three-expert combinations, and full models. Expert abbreviations: Id–Identity, Ad–Adaptive, At–Attention, SS–SpatioSemantic. Best results are in \textbf{bold}, second-best are \underline{underlined}.}
\label{tab:results_combined}
\centering
\footnotesize
\setlength{\tabcolsep}{3pt} 
\begin{tabular}{@{}l cccc cccc@{}}
\toprule
\multirow{2}{*}{\textbf{Config}} & 
\multicolumn{3}{c}{\textbf{METR-LA}} & \textbf{Time} & 
\multicolumn{3}{c}{\textbf{PEMS-BAY}} & \textbf{Time} \\
\cmidrule(lr){2-4} \cmidrule(lr){6-8}
& MAE & RMSE & MAPE & (s) & MAE & RMSE & MAPE & (s) \\
\midrule
\multicolumn{9}{c}{\textbf{Individual Experts}} \\
Id & 3.37 & 6.85 & 9.38\% & 1.46 & 1.73 & 3.85 & 4.01\% & 3.34 \\
Ad & 3.11 & 6.17 & 8.34\% & \underline{1.10} & \textbf{1.63} & \textbf{3.58} & \underline{3.65}\% & \underline{2.63} \\
At & 3.65 & 7.19 & 10.25\% & 1.63 & 1.88 & 4.13 & 4.24\% & 4.62 \\
SS & 3.13 & 6.33 & 8.65\% & \textbf{1.07} & \textbf{1.63} & \underline{3.61} & \underline{3.65}\% & \textbf{2.49} \\
\midrule
\multicolumn{9}{c}{\textbf{Two-Expert Combos}} \\
Id+Ad & \textbf{2.99} & \textbf{6.13} & \textbf{8.11}\% & 2.83 & 1.65 & 3.67 & \textbf{3.62}\% & 5.79 \\
Id+At & 3.47 & 7.15 & 9.83\% & 2.96 & 1.65 & 3.66 & \underline{3.65}\% & 8.76 \\
Id+SS & 3.23 & 6.47 & 8.71\% & 2.75 & 1.66 & 3.70 & 3.68\% & 6.41 \\
Ad+At & \underline{3.10} & 6.19 & 8.27\% & 2.61 & 1.93 & 4.23 & 4.33\% & 7.09 \\
Ad+SS & 3.11 & 6.37 & \underline{8.25}\% & 2.10 & 1.67 & 3.73 & 3.71\% & 4.95 \\
At+SS & 3.19 & 6.34 & 8.64\% & 2.61 & 1.70 & 3.83 & 3.76\% & 6.92 \\
\midrule
\multicolumn{9}{c}{\textbf{Three-Expert Combos}} \\
Id+Ad+SS & 3.14 & 6.17 & 8.47\% & 4.15 & \textbf{1.63} & 3.62 & 3.73\% & 9.81 \\
Id+At+SS & 3.27 & 6.57 & 8.92\% & 4.68 & 1.72 & 3.74 & 3.87\% & 11.76 \\
Ad+At+SS & 3.13 & 6.30 & 8.49\% & 3.65 & \underline{1.64} & 3.62 & \textbf{3.62}\% & 9.40 \\
\midrule
\multicolumn{9}{c}{\textbf{Full Models}} \\
TESTAM & 3.14 & 6.19 & 8.67\% & 4.73 & 1.72 & 3.71 & 3.91\% & 12.04 \\
TESTAM+ & \underline{3.10} & \underline{6.16} & 8.34\% & 6.04 & 1.65 & 3.65 & 3.73\% & 15.12 \\
\bottomrule
\end{tabular}
\end{table}

Among individual experts, the \textit{Adaptive Expert} (Ad) consistently outperforms others, achieving optimal results on PEMS-BAY with MAE of 1.63 and demonstrating near-state-of-the-art performance independently. The \textit{SpatioSemantic Expert} (SS) shows comparable accuracy with identical MAE but slightly elevated RMSE values. Conversely, the \textit{Attention Expert} (At) exhibits poor performance across both datasets, with MAE of 3.65 on METR-LA, validating our hypothesis that traditional attention mechanisms struggle with complex spatiotemporal dependencies in dense traffic networks.

The two-expert combinations validate our core principle of expert diversity over quantity. \textbf{Identity + \textit{Adaptive Expert} (Id+Ad) emerges as the optimal configuration}, achieving best performance on METR-LA while maintaining excellent results on PEMS-BAY. This pairing effectively balances structural simplicity through identity connections with the adaptability of learned spatial relationships. Combinations involving the \textit{Attention Expert} consistently underperformed, reinforcing the limitations of attention-based spatial modeling in traffic forecasting applications.

Three-expert configurations demonstrate diminishing returns, with Identity + Adaptive + SpatioSemantic experts (Id+Ad+SS) and Adaptive + Attention + SpatioSemantic experts (Ad+At+SS) achieving competitive but inferior performance compared to the simpler Id+Ad combination. This finding supports mixture-of-experts principles where \textbf{targeted specialization outperforms ensemble averaging}, particularly when computational efficiency is considered.

\paragraph{Computational Efficiency Trade-offs} Individual experts maintain low inference times (1.07--4.62 seconds), while expert combinations scale computational overhead proportionally. The Identity + \textit{Adaptive Expert} (Id+Ad) configuration reduces inference time by \textbf{53.1\%} on METR-LA and \textbf{61.7\%} on PEMS-BAY compared to TESTAM+ while achieving superior accuracy. The full TESTAM model exhibits the highest latency (15.12 seconds on PEMS-BAY), indicating that routing overhead significantly impacts scalability for real-time applications.

\paragraph{Critical Insights on Expert Effectiveness} Our analysis reveals that individual expert performance can surpass complex multi-expert ensembles. On PEMS-BAY, both Adaptive (1.63 MAE) and SpatioSemantic (1.63 MAE) experts individually outperform the original three-expert TESTAM (1.72 MAE), demonstrating that strategic specialization is more valuable than ensemble diversity. The \textit{SpatioSemantic Expert}'s ability to match the \textit{Adaptive Expert}'s performance while providing distinct spatial modeling capabilities validates our hybrid graph construction approach. This finding fundamentally challenges the assumption that more experts lead to better performance, suggesting that expert design should prioritize problem-specific specialization over architectural complexity.

\paragraph{Dataset-Specific Optimization} The results reveal clear dataset-specific patterns. METR-LA benefits from expert combination (Id+Ad achieving 2.99 MAE), while PEMS-BAY achieves optimal results with individual experts (Ad, SS both at 1.63 MAE). This suggests that traffic network characteristics and density influence optimal expert selection strategies, with denser networks (PEMS-BAY) requiring more focused spatial modeling approaches.
This structure improves readability and fits seamlessly into an academic paper. Let me know if you'd like to add bullet points, visual references, or citations.

\section{Conclusion \label{sec:conclusion}}

This work presents TESTAM+, an enhanced mixture-of-experts framework for traffic forecasting that addresses critical limitations in existing approaches through the introduction of a novel \textit{SpatioSemantic Expert} and analysis of expert composition strategies.
Our primary contribution lies in demonstrating that strategic expert selection fundamentally outperforms naive ensemble aggregation in spatiotemporal forecasting. Individual experts achieve remarkable effectiveness: the Adaptive and \textit{SpatioSemantic Expert} both achieve 1.63 MAE on PEMS-BAY, outperforming the original three-expert TESTAM (1.72 MAE). The Identity + \textit{Adaptive Expert} combination achieves state-of-the-art performance on METR-LA with 11.5\% MAE improvement over MegaCRN, while TESTAM+ consistently outperforms TESTAM with 1.3\% and 4.1\% MAE improvements across datasets. Simultaneously, these configurations reduce computational overhead by over 50\% compared to full multi-expert configurations. This finding challenges the conventional wisdom that more experts lead to better performance in mixture-of-experts architectures.
The \textit{SpatioSemantic Expert} successfully integrates physical road network topology with adaptive traffic patterns, matching the performance of the best individual expert while providing distinct spatial modeling capabilities. However, our ablation studies reveal that the complexity introduced by additional experts can harm rather than help performance when expert specialization is not properly aligned with the underlying problem structure.
\paragraph{Key limitations of existing approaches} TESTAM's three-expert design suffers from suboptimal expert combination strategies, where the routing mechanism fails to effectively leverage complementary strengths. Our analysis shows that the \textit{Attention Expert} consistently underperformed across datasets, suggesting that attention-based spatial modeling is fundamentally mismatched for dense traffic network dependencies. Furthermore, the computational overhead of multi-expert routing significantly impacts real-time deployment, with full TESTAM+ exhibiting 15.12-second inference times on PEMS-BAY.
\paragraph{Future research directions} (1) developing principled expert design methodologies that align architectural choices with domain-specific spatiotemporal characteristics, (2) investigating lightweight routing mechanisms that maintain expert diversity while reducing computational, and (3) exploring \textit{Adaptive Expert} activation strategies that dynamically adjust model complexity based on input characteristics.

\bibliographystyle{unsrt}
\bibliography{paper}

\end{document}